\PassOptionsToPackage{authoryear}{natbib}
\documentclass{article}
\usepackage{array}
\usepackage{multirow}
\usepackage{graphicx}
\usepackage{adjustbox}
\usepackage{array}
\usepackage{doi}
\usepackage{authblk}
\newcolumntype{L}[1]{>{\raggedright\let\newline\\\arraybackslash\hspace{0pt}}m{#1}}
\newcolumntype{C}[1]{>{\centering\let\newline\\\arraybackslash\hspace{0pt}}m{#1}}
\newcolumntype{R}[1]{>{\raggedleft\let\newline\\\arraybackslash\hspace{0pt}}m{#1}}
\newcolumntype{x}[1]{>{\centering\arraybackslash\hspace{0pt}}p{#1}}

\usepackage[utf8]{inputenc} 
\usepackage[T1]{fontenc}    
\usepackage{float}
\usepackage{hyperref}       
\usepackage{url}            
\usepackage{booktabs}       
\usepackage{amsfonts}       
\usepackage{nicefrac}       
\usepackage{microtype}      
\usepackage{amsmath}
\usepackage{tikz}
\usetikzlibrary{shapes,arrows}
\usetikzlibrary{arrows,calc}
\usepackage{enumerate}
\usepackage{enumitem}
\usepackage{pgfplots}
\usetikzlibrary{positioning}  
\hypersetup{hidelinks}

\title{Node-Level Financial Optimization in Demand Forecasting Through Dynamic Cost Asymmetry and Feedback Mechanism}

\author[3]{Alessandro Casadei}
\author[1]{Clemens Grupp}
\author[2]{Sreyoshi Bhaduri}
\author[3]{Lu Guo}
\author[3]{Wilson Fung}
\author[3]{Rohit Malshe}
\author[3]{Raj Ratan}
\author[3]{Ankush Pole}
\author[3]{Arkajit Rakshit}

\affil[1]{Amazon, Luxembourg, LU}
\affil[2]{Amazon, New York, NY}
\affil[3]{Amazon, Seattle, WA}

\date{}

\begin{document}

\maketitle

\begin{abstract}
This work introduces a methodology to adjust forecasts based on node-specific cost function asymmetry. The proposed model generates savings by dynamically incorporating the cost asymmetry into the forecasting error probability distribution to favor the least expensive scenario.
Savings are calculated and a self-regulation mechanism modulates the adjustments magnitude based on the observed savings, enabling the model to adapt to station-specific conditions and unmodeled factors such as calibration errors or shifting macroeconomic dynamics. Finally, empirical results demonstrate the model’s ability to achieve \$5.1M annual savings.
\end{abstract}

\section{Introduction}
Demand forecasting is traditionally evaluated based on its absolute error in predicting demand, facilitating capacity acquisition or decommitment decisions. However, the cost due to this error, referred as regret cost, can be different in case of overestimation or underestimation of demand. By using an error-regret cost function to capture this difference, forecasts can be adjusted to favor the least expensive type of error. This work introduces a model to apply an adjustment to a base forecast, strategically favoring either a positive (H: volume Heaviness, actual demand > forecast) or negative (L: volume Lightness, actual demand < forecast) forecasting error for each node (referred as station) in the network.

The known literature lacks of a framework that:
\begin{enumerate}[noitemsep]
    \item \textbf{Applies forecast adjustments at node-level} rather than country-level (Calzoni et al., 2023).
    \item \textbf{Leverages observed cost data} for prediction and evaluation, as existing approaches rely on an assumed, deterministic cost curve.
    \item \textbf{benefits from self-regulating capabilities} due to the feedback component.
\end{enumerate}
The significance of a node-level application based on observed data arises from the fact that nodes within a country may exhibit unique cost structures that evolve over time, influenced by factors such as:
\begin{enumerate}[noitemsep]
    \item \textbf{Macroeconomic factors:} Namely, proximity to countries with more expensive cost of labor. This is the case for nodes in Germany near the Swiss border face higher labor costs due to proximity to Switzerland, where wages are significantly higher. This makes H scenarios more cost-effective by reducing overstaffing costs.\\
    Another factor is labor market dynamics.
    This is the case in Mallorca (a tourism-reliant island in Spain), where competition with the tourism sector during high season drives up wages, favoring H scenarios compared to the rest of Spain. 
    \item \textbf{Internal factors:} Node-specific dynamics also affect regret costs. For instance, a positive work environment can lead to higher worker willingness to take overtime (covering H scenarios at a lower cost) or leverage voluntary time off (VTO), facing L scenarios at zero cost up to 5\% lightness.
\end{enumerate}
Observed costs at the node level capture these variations, providing a solid foundation for forecast adjustments across both geographic and temporal dimensions. 
The feedback component, instead, is necessary as the adjustment application is based on assumptions that may not always hold true in practice. This can lead to different savings outcomes over time. Accordingly, the forecasting adjustment is regulated based on these outcomes.
\tikzset{
    block/.style = {draw, rectangle, 
        minimum height=1cm, 
        minimum width=3cm},
    input/.style = {coordinate,node distance=1cm},
    output/.style = {coordinate,node distance=3cm},
    arrow/.style={draw, -latex,node distance=1cm},
    pinstyle/.style = {pin edge={latex-, black,node distance=2cm}}
}
\begin{figure}[h]
    \begin{center}
        \begin{tikzpicture}[auto, node distance=4cm,>=latex']
        \node [input, name=input] {};  
        \node [block, xshift=-2cm] (regret_cost) {Error \& Cost Data};
        \node [block, right of=regret_cost] (data_processing) {Data Processing};
        \node [block, right of=data_processing] (adjustment_application) {Adjustment Application};
        \node [output, right of=adjustment_application] (savings) {$S_s$ (Weekly Savings)};
        \node [block, below=0.5cm of adjustment_application] (adjustment_aggressiveness) {Modulation};

        \draw [->] (regret_cost) -- node {} (data_processing);
        \draw [->] (data_processing) -- node {} (adjustment_application);
        \draw [->] (adjustment_application) -- node {} (savings);
        \draw [->] (savings) |- node [above,pos=0.79] {} (adjustment_aggressiveness);
        \draw [->] (adjustment_aggressiveness) -- node {} (adjustment_application); 
        \end{tikzpicture}
    \end{center}
   \caption{Model's conceptual design for each node $s$ in the network.}\label{fig}
\end{figure}
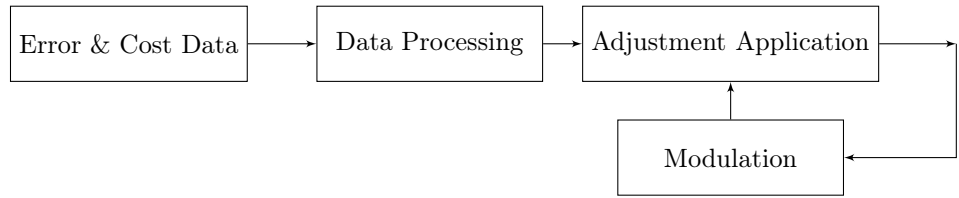

\textbf{Paper structure:} The remaining of the paper is organized as follow: the related works section provides a brief history of cost optimization approaches. Section \ref{sec:2} details the methodology, outlining the core components of the model. Section \ref{sec:3} exhibit the achieved financial results. Section \ref{sec:4} discusses the next steps and concludes the paper.

\textbf{Related works:} Cost optimization is historically a domain approached through LP (Linear Programming) or similar methods. Different approaches added stochasticity to LP problems to better reflect real scenarios.
Yu \& Li, 2000 added add a form of stochasticity to a traditional cost minimization operations research problem by identifying different scenarios and assign a probability to each. 
Santoso et al., 2004 includes stochasticity to LP network design problems by assigning a joint distribution to costs and capacity limits. 
Finally, Calzoni et al., 2023 presents the Palladio-NOSO framework which stands out due to its focus on cost optimal demand prediction, and, as such, it first applies a stochastical method (Palladio) and then applies a cost optimization (NOSO), hence framing the problem from an intrinsically stochastic perspective rather than adding stochasticity to an LP problem. The framework generates a stochastic demand forecast to then select the quantile that minimizes a cost function. 

\section{Methodology}
\label{sec:2}
The methodology centers on the WK-1 forecast in Amazon's last-mile operations, translating the forecasting error–regret cost relationship into a forecast adjustment strategy that minimizes costs. The approach unfolds in key steps:

First, the data processing component focuses on improving the prediction of regret cost based on forecasting error, with the aim of enhancing savings generation (as detailed in \textit{Table 1}).

Second, the modeling phase represents the regret cost function as two opposing linear relationships, with slopes equal to \(-\hat{CPP}_{Ls}\) (predicted Cost of Lightness Per Package) and \(\hat{CPP}_{Hs}\) (predicted Cost of Heaviness Per Package). This choice is based on the empirical fact that observed \(CPP_{Ls}\) and \(CPP_{Hs}\) values are strong predictors of future \(\hat{CPP}_{Ls}\) and \(\hat{CPP}_{Hs}\), achieving 84\% accuracy as measured by backtesting in 2024. This suggests that \(\hat{CPP}\) remains relatively stable for varying levels of forecasting error, hence can be modelled as constant.

Next, the probability distribution of the base forecast is adjusted so that the probabilities of lightness and heaviness, \(P(L)^*_s\) and \(P(H)^*_s = 1 - P(L)^*_s\) are inversely proportional to the respective asymmetry in the cost function. This is applied to increase the probability of the cheaper scenario to occur and, consequently, achieve savings. For example, if \(\hat{CPP}_{Ls} = 2 \cdot \hat{CPP}_{Hs}\), an adjustment is applied to the base forecast such that \(P(L)^*_s = \frac{1}{2} \cdot P(H)^*_s\).

Finally, the extent to which assumptions align with reality over time, as well as the accuracy of \(\hat{CPP}\) predictions, can influence savings generation positively or negatively. To account for this, a dynamic self-regulation mechanism is introduced, enabling greater forecast adjustments under favorable conditions and smaller adjustments otherwise.

\subsection{Data Processing}
The data processing component is further divided into two sections: time-weighting and noise reduction.

\subsubsection{Time-weighting}
Incorporating time-weighting has been shown to significantly enhance the modeling accuracy of causal relationships (Imbens, 2024), and this was also observed in the current study, which resulted in an annual savings of \$0.2M compared to models without time-weighting. The time-weighting is applied as follows: \\
Given a date-ordered dataset \(D'\) with \( n \) records, where each record \( i \) is associated with a date \( t_i \), we define a duplication factor \( w_i \) for each record \( i \) based on its recentness percentile such that:
\begin{equation}
w_i = \lceil k \cdot p_i \rceil \qquad k \in \mathbb{Z}^+ 
\end{equation}

where \( \lceil \cdot \rceil \) denotes the ceiling function, ensuring \( w_i \in \{1, 2, \dots, k\} \).
Note that \(k\) directly impacts both the granularity (number of bins) and magnitude (emphasis on more recent records) of the weighting. A value of\(k=10\) resulted in best cost prediction accuracy (84\%).

Each record \( i \) is repeated in \( w_i \) times to generate the new dataset \( D^* \):
\begin{equation}
D^* = \bigcup_{i=1}^{n} \{x_i\}^{w_i}
\end{equation}
where \( x_i \) is the \( i \)-th record in \( D' \), and \( \{x_i\}^{w_i} \) represents \( w_i \) copies of \( x_i \). The next steps are applied to \(D^*\). 

\subsubsection{Isolation of WK-1 Forecast-Driven Regret Cost through D-1 Noise Reduction} 
The regret cost \(C_{H}^\text{raw}, C_{L}^\text{raw}\) is the sum of several L and H cost metrics outlined in \textit{Appendix 1}. \(C_{H}^\text{raw}, C_{L}^\text{raw}\) incorporates costs due to other forecasting horizons, such as D-1, while the focus is on WK-1 horizon optimization. Hence, it is necessary to model the WK-1 forecasting error and its pure relationship with regret cost \(C_{H}, C_{L}\) by eliminating the cost (noise) introduced by the D-1 forecast \(N_{H}, N_{L}\). \\
Accordingly, \(N_{H}, N_{L}\) can be modelled under two assumptions:
\begin{enumerate}[itemsep=0pt]
    \item A cost within a \(\pm 5\%\) range (empirically acknowledged due to the limited time available for stations to acquire or decommit capacity).
    \item Stations adhere to flex-up/down processes triggered by the D-1 forecast.
\end{enumerate}

\begin{figure}[h]
\centering
\begin{tikzpicture}[
    node distance=0.8cm and 2cm, 
    every node/.style={draw, align=center, minimum height=0.8cm}, 
    ->, >=latex
]
\node (wk1_forecast) [draw=orange, minimum width=3cm] {WK-1 Last Mile Forecast (\({e}\))};
\node[left=of wk1_forecast] (hist_dem) {WK-1 Top Line Forecast};
\node[below=of wk1_forecast] (d1_forecast) {D-1 Last Mile Forecast};
\node[below=of wk1_forecast, xshift=-5cm] (d1_info) {Info available on D-1};
\node[right=of d1_forecast] (regret_cost) [draw=orange] {Regret Cost (\(C_{H}^\text{raw}, C_{L}^\text{raw}\))};

\draw[orange] (wk1_forecast) -- (regret_cost) node[midway, right, yshift=0.1cm, draw=none, fill=none, text=black] {\textit{\(C_H, C_L\)}};
\draw (wk1_forecast) -- (d1_forecast);
\draw (d1_info) -- (d1_forecast);
\draw (hist_dem) -- (wk1_forecast);
\draw (d1_forecast) -- (regret_cost);
\draw[dashed] (d1_forecast) -- (regret_cost) node[midway, below, yshift=-0.3cm, draw=none, fill=none] {\textit{Noise (\(N_H, N_L\))}};
\end{tikzpicture}
\vspace{-0.7cm} 
\caption{Causal DAG - Representation of the causal relationship between D-1 and WK-1 forecasts on regret cost.}
\end{figure}
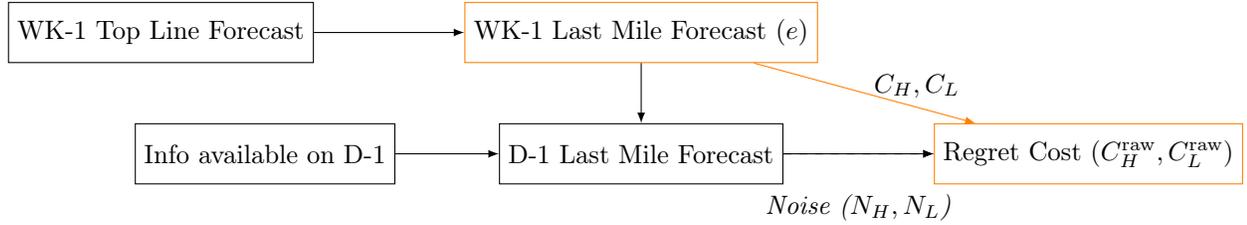

For parsimoniousness, the noise reduction is presented below only for H scenarios. Given:

\hspace{1cm} \(s: \text{station}\), \(t: \text{week number}\), \(o: \text{observed demand}\), \(f: \text{WK-1 forecast}\), \(f_2: \text{D-1 forecast}\),

\hspace{1cm} \(\epsilon = \frac{o - f}{f}: \quad \text{WK-1 forecasting error \%,}\)

\hspace{1cm} \(\varphi = \quad \frac{f_2 - f}{f}: \quad \text{D-1 \% forecast increase vs WK-1}\)

The utilization of the D-1 horizon, denoted as \( \mathcal{U}_{D-1} \), is defined as the percentage of the D-1 forecast that effectively reduces the WK-1 forecasting error. For a positive error \( \epsilon_{t,s} > 0 \) (denoted as H), this reduction occurs when the D-1 forecast exceeds the WK-1 forecast (\( \varphi > 0 \)). \\
If the increase in D-1 forecast relative to the WK-1 forecast \( \varphi \) is less than the WK-1 error \( e \), the entire D-1 increase is fully utilized, meaning D-1 reduces the regret cost by \( \varphi \). However, if \( \varphi > e \), only the amount equal to \( e \) is utilized, and the remaining portion (\( \varphi - \mathcal{U}_{D-1} \)) represents the non-utilized portion of the D-1 increase. If \(\varphi < 0\), D-1 is entirely increasing the WK-1 error, hence \(\mathcal{U}_{D-1} = 0\). Formally, \(
\forall \, {t,s} \text{ and } \epsilon > 0\):
\[
\mathcal{U}_{D-1} = 
\begin{cases} 
\min(\epsilon, \varphi), & \text{if } \varphi \geq 0, \\
0, & \text{if } \varphi < 0.
\end{cases}
\]
The D-1 utilization directly translates into the regret cost nuisance generated by the D-1 forecast when multiplied by the regret cost \( C_{H}^\text{raw} \), with \( \mathcal{U}_{D-1} \) capped at 5\% as per assumption 1. Note that, as we are considering only H scenarios, the cost generated by the D-1 forecast \( N_{H} \) is negative (savings are generated) when the increase versus the WK-1 forecast \( \varphi \) is positive, while costs are positive when \( \varphi \) is negative. The opposite is true for \( N_{L} \). Formally:
\[
N_{H} = 
\begin{cases} 
-\min(\mathcal{U}_{D-1}, 0.05) \cdot C_{H}^\text{raw}, & \text{if } \varphi \geq 0, \\
\min(|\varphi| - \mathcal{U}_{D-1}, 0.05) \cdot C_{H}^\text{raw}, & \text{if } \varphi < 0.
\end{cases}
\]
\[
N_{L} = 
\begin{cases} 
\min(\varphi - \mathcal{U}_{D-1}, 0.05) \cdot C_{L}^\text{raw}, & \text{if } \varphi \geq 0, \\
-\min(\mathcal{U}_{D-1}, 0.05) \cdot C_{L}^\text{raw}, & \text{if } \varphi < 0.
\end{cases}
\]
The cost nuisance due to the D-1 forecast is then subtracted from the regret cost to obtain the regret cost purely due to the WK-1 forecast \(C_H, C_L\):
\[
C_{H} = C_{H}^\text{raw} - N_{H}, \quad C_{L} = C_{L}^\text{raw} - N_{L}.
\]
This adjustment ensures accurate attribution of costs to specific forecast horizons, with future analyses considering all horizons for enhanced precision.

The regret cost function is now modeled using \(C_H\) and \(C_L\) for each station \(s\) and week \(t\), serving as the basis for optimization via forecast adjustment. This is achieved by calculating the predicted Costs Per Package \(\hat{CPP}_{Ls}\) and \(\hat{CPP}_{Hs}\), representing the regret cost driven by one package of L and H, respectively. \\
For each station, \( \hat{CPP}_{H} \) and \( \hat{CPP}_{L} \) are the averages of \( CPP_{H} \) and \( CPP_{L} \) over the dataset \( D^* \), only for the H and L scenarios, respectively. This approach excludes cases with mismatching costs and scenarios (e.g., it excludes \( C_H \) when \( e < 0 \)) to ensure that regret cost that cannot be due to WK-1 forecasting error is not captured. Formally, \(\forall {s}:\)
\begin{equation}
\hat{CPP}_{Hs} = \frac{\sum_{t=1}^{d_s} C_{H_{s,t}} \mathbb{I}(e_{s,t} > 0)}{\sum_{t=1}^{d_s} e_{s,t} \mathbb{I}(e_{s,t} > 0)}.
\end{equation}
\begin{equation}
\hat{CPP}_{Ls} = \frac{\sum_{t=1}^{d_s} C_{L_{s,t}} \mathbb{I}(e_{s,t} < 0)}{\sum_{t=1}^{d_s} e_{s,t} \mathbb{I}(e_{s,t} < 0)}.
\end{equation}

\subsection{Regret cost function definition}

Empirical results show that \( \hat{CPP}_{H} \) and \( \hat{CPP}_{L} \) are reliable predictors of the observed \({CPP}_{H} \) and \({CPP}_{L} \), with a 84\% prediction accuracy. This suggests that \(CPP_H\) and \(CPP_L\) are constant over different intervals of \(e\), meaning that the regret cost function remains piecewise constant with respect to \(e\).
Accordingly, the cost curve for a given week \(t\) and station \(s\) is modeled as two opposing lines with constant slopes \(-\hat{CPP}_{L}\) and \(\hat{CPP}_{H}\):
\begin{equation}
\hat{C} =
\begin{cases} 
-\hat{CPP_{L}} \cdot e & \text{if } e \leq 0, \\
\hat{CPP_{H}} \cdot e & \text{if } e > 0.
\end{cases}
\end{equation}
\begin{figure}[h]
\centering
\begin{tikzpicture}
    \begin{axis}[
        width=8cm,          
        height=5cm,         
        grid=both,
        grid style={dotted,gray!50},
        xlabel={Forecasting Error $e$},
        ylabel={Regret Cost $\hat{C}$},
        axis lines=middle,
        enlargelimits={abs=0.5cm},  
        legend pos=outer north east,
        every axis plot/.append style={thick},
        label style={font=\small},   
        legend style={font=\small},  
        title style={font=\small},   
        xlabel style={xshift=0.5cm}
    ]
        \addplot[blue, domain=-5:0, samples=100] {-1 * x};
        \addlegendentry{$\hat{C}$ for $e \leq 0$};

        \addplot[red, domain=0:5, samples=100] {5 * x};
        \addlegendentry{$\hat{C}$ for $e > 0$};
    \end{axis}
\end{tikzpicture}
\caption{Regret cost function}
\label{fig:cost_function}
\end{figure}
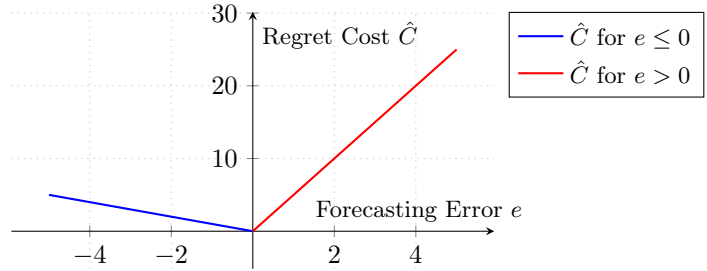

\subsection{Adjustment of Forecasting Error Probability to Reflect Regret Cost Asymmetry}
In order to achieve the optimal probability \(P(L)^*_s\), it is first necessary to define the base forecast and the initial probability \(P(L)\). Once defined, the required forecast adjustment can be calculated to transform the base forecast into an adjusted forecast with \(P(L) = P(L_s)^*\).

The base forecast is defined as a symmetric, zero-mean Gaussian distribution, where \(P(L) = P(H) = 0.5\), reflecting an absence of intentional adjustments toward either scenario in Amazon's last-mile forecasting process up to this point. Note that this representation assumes perfect calibration.
At this stage, an adjustment \(\Delta\) has to be applied to this distribution in order to set \(P(L) = P(L)^*_s\).
The adjusted forecast corresponds to an optimal horizontal shift \(\Delta\) of the distribution (see \textit{figure 4}).
\begin{itemize}
    \item Forecasting error \(e\): \(o-(f+\Delta)\), difference between observed demand and base forecast (if \(\Delta=0\)), adjusted forecast otherwise
    \item \(\sigma\): standard deviation of the distribution of \(e\) for station \(s\) at week \(t\)
\end{itemize}

We want to minimize the total expected cost defined as the error distribution of an adjusted forecast \(P(e)\) (as per \textit{figure 4}) multiplied by the regret cost associated to \(e\) (as per \textit{figure 3}). For a given week \(t\) and station \(s\), this translates into:
\begin{equation}
\Delta^* = \min_\Delta \int_{-\infty}^\infty \hat{C}(e) \cdot P(e) \, de.
\end{equation}
Considering that positive values of \(\Delta\) lead to a negative mean error \(e\), the probability distribution of the forecasting error \( e \) under the adjusted forecast is:
\begin{equation}
P(e) = \frac{1}{\sigma \sqrt{2\pi}} \exp\left(-\frac{(e + \Delta)^2}{2\sigma^2}\right)
\end{equation}

\begin{figure}[h]
    \centering
\begin{tikzpicture}
    \begin{axis}[
        axis lines = middle, 
        enlargelimits,
        xlabel = \( e \),
        ylabel = \( P(e) \),
        domain=-5:5,
        samples=100,
        width=10cm,
        height=7cm,
        grid=major,
        ticklabel style={font=\tiny},
        legend style={font=\tiny},
    ]
    
    \addplot[orange, thick, smooth] {1/sqrt(2*pi)*exp(-x^2/2)};
    \addlegendentry{Mean = 0, Std = \(\sigma\)}

    \addplot[green, thick, smooth] {1/sqrt(2*pi)*exp(-(x+1)^2/2)};
    \addplot [gray!20,opacity=0.6, 
        domain=-5:0,
        smooth,
        fill
    ] {1/sqrt(2*pi)*exp(-(x+1)^2/2)} \closedcycle;
    \addlegendentry{Mean = \(\Delta^*\), Std = \(\sigma\)}

    \addplot[orange, opacity=0.3, domain=-5:0] {1/sqrt(2*pi)*exp(-x^2/2)} \closedcycle;

    \addplot[green, opacity=0.3, domain=-5:0] {1/sqrt(2*pi)*exp(-(x+1)^2/2)} \closedcycle;

    \draw[dashed, red] (axis cs:-1,0) -- (axis cs:-1,0.4);

    \draw[dashed, red] (axis cs:0,0.4) -- (axis cs:-1,0.4);

    \node at (axis cs:0,-0.05) [anchor=north] {Mean = 0};
    \node at (axis cs:-1,-0.05) [anchor=north] {Mean = -1};

    \node at (axis cs:2.8,0.2) {\(P(L)\) = 0.5};
    \node at (axis cs:-4,0.2) {\(P(L)=P(L)^*\)};

    \draw[dashed, red, thick] (axis cs:0,0) -- (axis cs:-1,0);
    \node at (axis cs:-1,0.00) [anchor=north] {\small \(\Delta^*\)};

    \end{axis}
\end{tikzpicture}
    \caption{Applying a positive forecast adjustment \(\Delta\) based on \(P(L)^*_s\) (red curve) starting from the error distribution of the base forecast (blue curve).}
    \label{fig:tweak-optimal}
\end{figure}
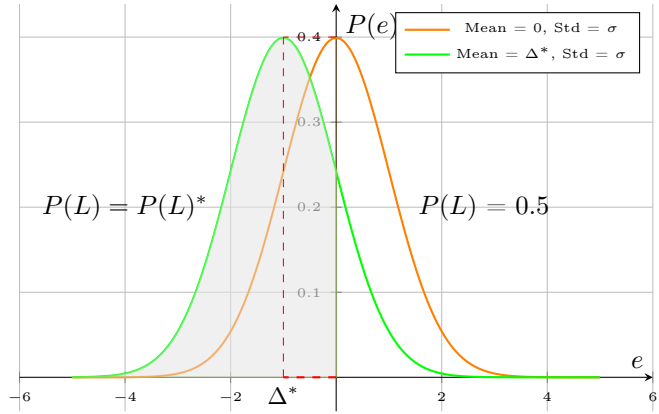

Accordingly, the expected cost of lightness \(\mathbb{E}[\hat{C_L}]\), i.e. cost occurring when (\(e<=0\)) is given by: 
\begin{equation}
\mathbb{E}[\hat{C_L}] = \int_{-\infty}^0 -\hat{CPP}_{L} \cdot e \cdot \frac{1}{\sigma \sqrt{2\pi}} \exp\left(-\frac{(e + \Delta)^2}{2\sigma^2}\right) \, de
\end{equation}
Similarly, the expected cost of heaviness \(\mathbb{E}[\hat{C_H}]\), i.e. cost occurring when (\(e>0\)) is given by:
\begin{equation}
\mathbb{E}[\hat{C_H}] = \int_{0}^\infty \hat{CPP}_{H} \cdot e \cdot \frac{1}{\sigma \sqrt{2\pi}} \exp\left(-\frac{(e + \Delta)^2}{2\sigma^2}\right) \, de
\end{equation}
We want to find \(\Delta^*\) as the value of \(\Delta\) that minimizes the total expected cost \(\mathbb{E}[\hat{C}]=\mathbb{E}[\hat{C_L}]+\mathbb{E}[\hat{C_H}]\). By replacing equations 8 and 9 in equation 6 the optimization problem becomes:,
\begin{equation}
\Delta^* = \min_\Delta \left[
\begin{aligned}
&\int_{-\infty}^0 -\hat{CPP}_{L} \cdot e \cdot \frac{1}{\sigma \sqrt{2\pi}} \exp\left(-\frac{(e + \Delta)^2}{2\sigma^2}\right) \, de \\
&+ \int_{0}^\infty \hat{CPP}_{H} \cdot e \cdot \frac{1}{\sigma \sqrt{2\pi}} \exp\left(-\frac{(e + \Delta)^2}{2\sigma^2}\right) \, de
\end{aligned}
\right]
\end{equation}

\begin{figure}[h]
    \centering
\begin{tikzpicture}
    \begin{axis}[
        axis lines = middle,
        enlargelimits,
        xlabel = {$e$},
        domain = -5:5,
        samples = 200,
        width = 12cm,
        height = 8cm,
        grid = major,
        tick label style = {font=\small},
        label style = {font=\small},
        title style = {font=\small},
    ]
        \addplot[domain=-5:0, samples=100, thick, blue] {-x * (1/(sqrt(2*pi))) * exp(-((x+1)^2)/2)};
        \addlegendentry{for $e \leq 0$}
        \addplot[domain=0:5, samples=100, thick, red] {5 * x * (1/(sqrt(2*pi))) * exp(-((x+1)^2)/2)};
        \addlegendentry{for $e > 0$}
        
        \addplot[domain=-5:0, samples=100, fill=blue!20, opacity=0.6] 
            {-x * (1/(sqrt(2*pi))) * exp(-((x+1)^2)/2)} \closedcycle;
        \addplot[domain=0:5, samples=100, fill=red!20, opacity=0.6] 
            {5 * x * (1/(sqrt(2*pi))) * exp(-((x+1)^2)/2)} \closedcycle;
    \node at (axis cs:-3.8,0.2) {\(\mathbb{E}[\hat{C_L}]\)}; 
    \node at (axis cs:2.2,0.2) {\(\mathbb{E}[\hat{C_H}]\)};

    \end{axis}
\end{tikzpicture}
\caption{Minimum \(\mathbb{E}[\hat{C}]=\mathbb{E}[\hat{C_L}]+\mathbb{E}[\hat{C_H}]\) with \(\Delta=\Delta^*\)}
\end{figure}
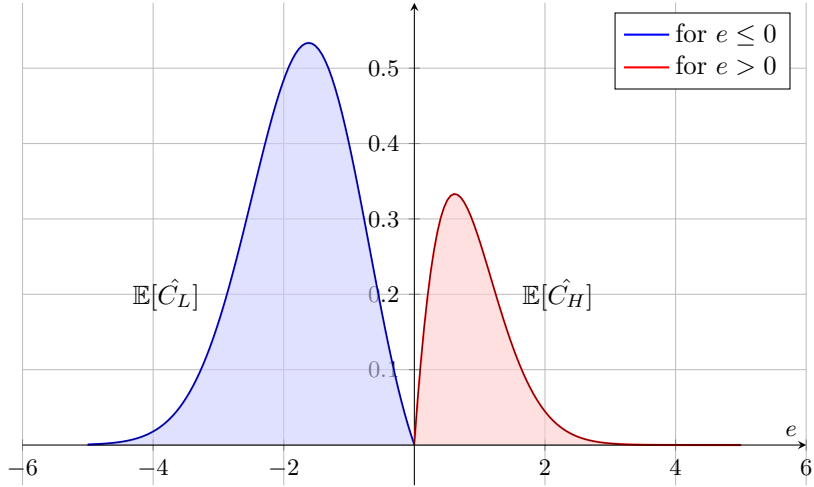

\subsection{Application of dynamic self-regulation}

The self-regulating aspect of the framework relies on evaluating the model's performance, measured by the weekly incremental cost \({C}\) generated by the adjustment \(\Delta\) compared to the cost incurred if we were not to apply any adjustment (\(\Delta=0\)).\\
Before calculating the cost generation we introduce the adjustment utilization \({\cal U}\), which identifies the portion of adjustment that reduces the error and, in turn, generates savings. Conversely, \(1-{\cal U}\) is the non-utilized adjustment, which increases the forecasting error and, in turn, generates costs. Equation (11) ensures that utilization is between 0\% and 100\% without altering the rigorousness of the cost calculation, as in (12) adjustments that would lead to \({\cal U} < 0\%\) entirely account for cost generation, while adjustments that would lead to \({\cal U} > 100\%\) entirely account for savings generation. Equations (11) and (12) are defined only for positive adjustments for parsimoniousness.
\(\forall \Delta > 0:\) \\
\(o: \text{observed demand}\), \(f: \text{base forecast (i.e. unadjusted)}\)
\begin{equation}
\mathcal{U} = \min\left(1, \frac{\max(0, o - f)}{\Delta}\right)
\end{equation}
Next, the non-utilized adjustment (\(\Delta \cdot (1-{\cal U})\)) multiplied by the cost per package incurred (\(CPP_L\) for positive \(\Delta\)) determines the cost generated. Similarly, the savings are calculated as the utilized adjustment (\(\Delta \cdot {\cal U}\)) multiplied by the avoided cost per package \(CPP_H\). The difference between these two components yields the total cost generated \({C}\):
\begin{equation}
\begin{split}
C &= \Delta \cdot (1 - \mathcal{U}) \cdot CPP_L - \Delta \cdot \mathcal{U} \cdot CPP_H \\
  &= \Delta \cdot [(1 - \mathcal{U}) \cdot CPP_L - \mathcal{U} \cdot CPP_H]
\end{split}
\end{equation}
Savings are generated when \((1 - \mathcal{U}) \cdot CPP_L < \mathcal{U} \cdot CPP_H\), otherwise costs are generated, with the magnitude of the savings/cost is scaled by a factor \(\Delta\).

By comparing the observed cost \(C\) with the expected cost \(\mathbb{E}[\hat{C}]\), we can gain insights to refine the model through a feedback mechanism. The aim is to apply a larger adjustment when the explainable difference between observed cost and expected cost (\(\epsilon_{calibration}, \epsilon_{\hat{CPP}}\)) is consistently negative (indicating more savings can be achieved) and a smaller one in the opposite case. \(C\) can be rewritten as:
\begin{equation}
   C = \mathbb{E}[\hat{C}] + \epsilon_{calibration} + \epsilon_{\hat{CPP}} + \epsilon_{unexp.}
\end{equation}
\begin{itemize}[itemsep=0pt, parsep=0pt]
\setlength{\itemsep}{\baselineskip}
    \item \textbf{\(\epsilon_{calibration}\)}: being the calibration error defined as the deviation from the assumption that the base forecast distribution is perfectly calibrated, with a mean equal to 0 (see \textit{Figure 4}). This is captured, for \(\Delta>0\), by \(\max(0, o - f)\) in (11), representing the observed equivalent of the base forecast for \(e > 0\). A sufficiently large sample of \(o - f)\) forms a probability distribution of the observed error, generalized for any values of $\Delta$. The mean of this distribution reflects the calibration error. The cost associated with the calibration error is defined as the difference between the expected cost calculated with the observed mean \(\mathbb{E}[o - f]\) and the expected cost calculated under the assumption of a mean equal to 0:
    \begin{equation}
    \begin{aligned}
    \epsilon_{calibration} = &\int_{-\infty}^0 \hat{C}(e) \cdot \frac{1}{\sigma \sqrt{2\pi}} \exp\left(-\frac{\left(e - \mathbb{E}[o - f] + \Delta\right)^2}{2\sigma^2}\right)\, de \\
    &- \int_{-\infty}^0 \hat{C}(e) \cdot \frac{1}{\sigma \sqrt{2\pi}} \exp\left(-\frac{(e + \Delta)^2}{2\sigma^2}\right) \, de
    \end{aligned}
    \end{equation}

    To adjust for this error in future predictions, the term \(\mathbb{E}[o - f]\) is integrated in equation 10.
    
    \item \textbf{\(\epsilon_{\hat{CPP}}\)}: being the cost prediction error due to the model's inaccuracy in predicting \(CPP_L, CPP_H\). The error is proportional to the difference between observed values \(CPP_{L}, CPP_{H}\) and predicted values \(\hat{CPP}_{L}, \hat{CPP}_{H}\). This can be driven by imperfect accuracy in time weighting and noise reduction. \(\epsilon_{\hat{CPP}}\) is defined as the difference between (10) with \(CPP_{L}, CPP_{H}\) replacing \(\hat{CPP}_{L}, \hat{CPP}_{H}\) and (10) itself. For each week \(t\) and station \(s\), this can be written as:
    \begin{equation}
    \epsilon_{\hat{CPP}}=\int_{-\infty}^0 -(CPP_L-\hat{CPP}_{L}) \cdot e \cdot P(e), de
    + \int_{0}^\infty (CPP_H-\hat{CPP}_{H}) \cdot e \cdot P(e), de
    \end{equation}
    The average $CPP_L-\hat{CPP}_{L}$ and $CPP_H-\hat{CPP}_{H}$ are named \(\overline{CPP_L}\), \(\overline{CPP_H}\). The average prediction errors are then added back in (10) so that future expected prediction errors are set to 0.

    \item \textbf{\(\epsilon_{unexp.}\)}: being any error not captured by \(\epsilon_{calibration}\) or \(\epsilon_{\hat{CPP}}\). This can be due to random noise, deviations in the regret cost function from linearity or deviations in the error distribution from a Gaussian profile.
\end{itemize}

Note that all the errors above can be negative, meaning they can translate into more savings than expected. This occurs when \( CPP_{Hs,t} - CPP_{Ls,t} > \hat{CPP}_{Hs,t} - \hat{CPP}_{Ls,t} \), leading to a higher cost asymmetry than predicted. For calibration errors' impact on savings, see Calzoni et al., 2023, section 3.2.

In light of the additional saving opportunities achievable by reducing the explainable errros \(\epsilon_{calibration}\) and \(\epsilon_{\hat{CPP}}\), we re-frame the optimization problem in equation 10 by integrating $\mathbb{E}[o - f]$, $\overline{CPP_L}$ and $\overline{CPP_H}$:
\begin{equation}
\Delta^* = \min_\Delta \left[ 
\begin{aligned}
&\int_{-\infty}^0 -(\hat{CPP}_{L} + \overline{CPP_L}) \cdot e \cdot \frac{1}{\sigma \sqrt{2\pi}} \exp\left(-\frac{\left(e - \mathbb{E}[o - f] + \Delta\right)^2}{2\sigma^2}\right) \, de \\
&+ \int_{0}^\infty (\hat{CPP}_{H} + \overline{CPP_H}) \cdot e \cdot \frac{1}{\sigma \sqrt{2\pi}} \exp\left(-\frac{\left(e - \mathbb{E}[o - f] + \Delta\right)^2}{2\sigma^2}\right) \, de
\end{aligned}
\right]
\end{equation}

\section{Financial results}
\label{sec:3}
In our analysis of the financial results, we began by calculating the total achievable savings for 2024, assuming a perfect adjustment that would lead to zero errors (\(e_{s,t} = 0\)) for all \(s,t\). The maximum achievable savings for the EU last mile alone were estimated to be \$25.7M. To assess the effectiveness of our model, we backtested the model to determine the portion of the maximum savings that it could attain. If applied in 2024, the model was expected to attain \$5.1M, 19.8\% of the maximum achievable savings. \\
Additionally, a test was conducted to examine whether data processing steps adequately improved the model's predictive power. Details on the annual cost generation with and without time-weighting and noise reduction are summarized in \textit{Table 1}. \\

\begin{table}[H]
\caption{Time weighting and noise reduction tests. EU Last Mile Network, Full Year 2024.}
\begin{tabular}{C{3cm}C{4cm}C{3.5cm}C{1.5cm}}
\toprule
\textbf{Time Weighting} & \textbf{Noise Reduction} & \textbf{Annual Cost Generated} & \textbf{Avg \({\cal U}\)} \\
\midrule
Y & Y & -\$5.1M & 39.74\% \\
Y & N & -\$4.9M & 39.91\% \\
N & Y & -\$4.9M & 39.77\% \\
N & N & -\$4.7M & 39.91\% \\
\bottomrule
\end{tabular}
\end{table}

\section{Future work}\label{sec:4}
\textbf{Implementation:} the implementation plan begins with a pilot in January 2025 in Italy, in collaboration with the IT S\&OP team, to evaluate the savings generated.\\
\textbf{Further development:} There is significant potential to broaden the framework’s scope by extending it beyond last mile and WK-1 forecasting horizon. Ideally, adjustments should be tailored to each horizon based on the capacity decisions they influence.\\
\textbf{Long-term:} Finally, the proposed framework should evolve to incorporate factors beyond efficiency when suggesting adjustments. According to the literature, efficiency, safety, and delivery timeliness are consistently identified as key pillars of a logistics network (Ekici, Kabak, \& Ülengin, 2019; Dudek, 2020). Integrating predictions of forecast-driven safety and timeliness impacts will enable a more comprehensive approach, as well as produce forecasts customized to align with the strategic prioritization of these pillars.
Prioritizing safety and delivery timeliness over efficiency should result in positive adjustments, ensuring additional capacity for the same demand. This approach reduces network strain, thereby enhancing safety and improving delivery performance.

\section*{Acknowledgments}
I would like to express my sincere gratitude to Clemens Grupp for his continuous feedback, guidance and support throughout the development of this project and the writing of this paper. His expertise has been invaluable at every stage of this work.



\section*{Appendix 1}
Regret cost composition. \\
\textbf{Acronyms}: \\
AMZL - Amazon's last mile \\
UTR - Under the Roof, warehouse internal operations in last mile \\
OTR - On the Road, delivering operations in last mile \\
CPH - Cost per Hour \\
TPH - Throughput per Hour (UTR efficiency measure) \\
DPPH - Deliveries per Paid Hour (OTR efficiency measure) \\
3Ps - Third-party logistics. Companies AMZL relies on when not capable to deliver the entirety of the volume

\begin{table}[H]
\centering
\caption{Regret Cost Metrics}
\renewcommand{\arraystretch}{1.5}
\setlength{\tabcolsep}{5pt}
\begin{adjustbox}{max width=\textwidth}
\begin{tabular}{|L{3cm}|L{3cm}|L{3cm}|L{5.5cm}|L{5.5cm}|}
\hline
\textbf{Heaviness/ Lightness Cost} & \textbf{Metric Group} & {\textbf{Metric Name}} & \textbf{Definition} & \textbf{Logic} \\ \hline
\multirow{8}{3cm}{Volume Lightness Total Cost} &
\multirow{5}{3cm}{OTR - Total Volume Lightness Cost} 
& OTR Cancellation Costs 
& Costs resulting from Payments to DSPs for cancelling promised routes. 
& Directly sourced from payments data. \\ \cline{3-5}
&
& OTR - Overstaffing 
& Costs resulting from OTR overstaffing vs forecast, resulting in spreading dispatched parcels among more routes than needed.
& Overstaffing hours * On Zone CPH \\ \cline{2-5}
&
{UTR - Total Volume Lightness Cost} 
& UTR - Overstaffing 
& TPH impact of actual volume vs forecast lower by more than -5\%. \newline Note: A decrease between 0\% and 5\% has no cost impact as it can be matched by a similar decrease in capacity by leveraging voluntary time off.
& Overstaffing hours * UTR cost per hours. \\ \hline
\multirow{8}{3cm}{Volume Heaviness Total Cost} &
{OTR - Total Volume Heaviness Cost} 
& OTR - Density due to Caps 
& DPPH lost opportunity driven by lower OTR density due to constrained capacity (named "capping"). 
& Loss due to AMZL caps / or\_disp\_ship\_amzl  / coefficient * OTR CPP *(1- fixed\_cpp\_over\_variable\_pctg) * or\_disp\_ship\_amzl \\ \cline{2-5}
&
{UTR - Total Volume Heaviness} 
& UTR - TPH Flex up
& Cost of overtime. 
& Flex up hours * overtime rate \\ \cline{2-5}
& \multirow{2}{3cm}{E2E - Amazon Impact} & 3P vs AMZL Cost due to caps & Difference in price between 3Ps and AMZL applied to volume delivered by 3Ps due to caps. & \% of loss due to caps * AMZL vs 3P * contributions \\ \cline{3-5}
& & UTR - TPH Flex up & Cost of overtime & Flex up hours * overtime rate  \\ \cline{2-5}
& \multicolumn{2}{|c|}{Speed Costs} & Future approximated revenue loss due to not meeting customers' promised delivery date. & Late delivered items * decreased loyalty cost 
\\ \hline
\end{tabular}
\end{adjustbox}
\end{table}

\section*{Appendix 2}
To establish the validity of Equation 10 and justify the search for $\Delta^*$, we first have to demonstrate that $\hat{C}$ is finite. If $\hat{C}$ were infinite, the concept of finding $\Delta^*$ as the $\Delta$ associated with the lowest cost would be meaningless. We approach this proof by examining the behavior of the integral over \(e>0\) and for a fixed value of $\Delta$ for simplicity.

As $e \to \infty$, the Gaussian term $\frac{1}{\sigma\sqrt{2\pi}}e^{-\frac{(e+\Delta)^2}{2\sigma^2}}$ exhibits exponential decay. This rapid decrease dominates the linear growth of the term $\hat{CPP}_{H} \cdot e$, ensuring the convergence of the integral. Formally:
\begin{equation}
\lim_{e \to \infty} \hat{CPP}_{H} \cdot e \cdot \frac{1}{\sigma \sqrt{2\pi}} e^{-\frac{(e + \Delta)^2}{2\sigma^2}} = 0.
\end{equation}
It's not complicated to prove the same for the integral over \(e \leq 0\) integral. As both integrals converge due to the fast exponential decay of the Gaussian function, the overall integral is finite. We can conclude that \(\hat{C}\) is also finite for different values of \(\Delta\) as \(\Delta\) is not affecting the convergence.

\small
\nocite{*}
\bibliographystyle{apalike}
\bibliography{refrence}

\end{document}